\documentclass{bmvc2k}
\usepackage{multirow}
\usepackage{booktabs}
\usepackage{graphicx}
\usepackage{amssymb}
\usepackage{subfigure}

\title{Unsupervised Domain Adaptation for Spatio-Temporal Action Localization}

\addauthor{Nakul Agarwal}{nagarwal@honda-ri.com}{12}
\addauthor{Yi-Ting Chen}{ychen@honda-ri.com}{2}
\addauthor{Behzad Dariush}{bdariush@honda-ri.com}{2}
\addauthor{Ming-Hsuan Yang}{mhyang@ucmerced.edu}{13}

\addinstitution{
 University of California, Merced
}
\addinstitution{
 Honda Research Institute USA
}
\addinstitution{
 Google Research
}
\runninghead{Agarwal,Chen,Dariush,Yang}{Domain Adaptation for Action Localization}

\begin{document}

\maketitle

\begin{abstract}

   %
   %
   %
   %
   %
   
Spatio-temporal action localization is an important problem in computer vision that involves detecting where and when activities occur, and therefore requires modeling of both spatial and temporal features. 
This problem is typically formulated in the context of supervised learning, where the learned classifiers operate on the premise that both training and test data are sampled from the same underlying distribution. 
However, this assumption does not hold when there is a significant domain shift, leading to poor generalization performance on the test data. 
To address this, we focus on the hard and novel task of generalizing training models to test samples without access to any labels from the latter for spatio-temporal action localization by proposing an end-to-end unsupervised domain adaptation algorithm. 
We 
extend the state-of-the-art object detection framework to localize and classify actions. 
In order to minimize the domain shift, three domain adaptation modules at image level (temporal and spatial) and instance level (temporal) are designed and integrated.
%
We design a new experimental setup and evaluate the proposed method and different adaptation modules on the UCF-Sports, UCF-101 and JHMDB benchmark datasets.
We show that significant performance gain can be achieved when spatial and temporal features are adapted separately, or jointly for the most effective results.
%
\end{abstract}
\vspace{-8pt}
\section{Introduction}
\label{intro}
Recently, there has been a significant interest in tackling the spatio-temporal human action localization problem due to its importance in many applications. 
Based on the recent benchmark datasets~\cite{soomro2012ucf101,weinzaepfel2016towards,gu2018ava} and temporal neural networks~\cite{tran2015learning,carreira2017quo}, numerous algorithms for spatio-temporal action localization have been proposed.
Although significant advances have been made, existing algorithms generally require a large-scale labeled dataset for supervised learning which i) is non-trivial and not scalable because annotating bounding boxes is expensive and time consuming and ii) do not generalize well when there is a significant domain shift between the underlying distributions in the training and test datasets. 
This domain shift can be caused by difference in scenarios, lighting conditions or image appearance. 
In case of videos, the variation in the progression of activity over time can also cause domain shift. 
Such domain discrepancy causes unfavorable model generalization. 
%
%
 %
%
%
%
\vspace{-7pt}

To address problems associated with domain shift, various domain adaptation algorithms
have been proposed.
Nevertheless, the majority of existing methods focus on images rather than video, catering to problems associated with image classification~\cite{saenko2010adapting,tzeng2014deep,long2015learning,ganin2015unsupervised}, semantic segmentation~\cite{sankaranarayanan2018learning,zou2018unsupervised,tsai2018learning} and object detection~\cite{chen2018domain,saito2019strong}. 
%
%
%
%
%
%
The ones that do focus on video action understanding can be divided into three categories: whole-clip action recognition, action segmentation, and spatio-temporal action localization. 
Some progress has been made in this field but only for the former two categories~\cite{jamal2018deep,chen2019temporal,pan2019adversarial,chen2020action1,chen2020action2}, while the latter category remains unattended.
%
%
%
Therefore, it is of great interest to develop algorithms for adapting spatio-temporal action localization models to a new domain.
%

%
\vspace{-10pt}

In this work, we focus on the hard problem of generalizing training models to target samples without access to any form of target labels for spatio-temporal action localization by proposing an end-to-end trainable unsupervised domain adaptation framework based on the Faster R-CNN ~\cite{ren2015faster} algorithm.  
%
%
To reduce the impact of domain shift, we design and integrate adaptation modules to jointly align both spatial and temporal features. 
Specifically, three adaptation modules are proposed: i) for aligning temporal features at the image level, ii) for aligning temporal features at the instance level and iii) for aligning spatial features at the image level. 
In each module, we train a domain classifier and employ adversarial training to learn domain-invariant features. 
For aligning the temporal features, both instance-level as well as image-level adaptation are considered. 
While the former focuses on the actor/action dynamics, the latter incorporates global scene features as context for action classification, which has shown to be effective~\cite{sun2018actor}. 
%
\vspace{-12pt}

Existing video action understanding datasets are not designed for developing and evaluating domain adaptation algorithms in the context of spatio-temporal action localization. 
To validate the proposed algorithm, we design new experimental settings. 
We first focus on the scenario of adapting to large scale data using a smaller annotated domain to show that we can harvest more from existing resources. 
We then provide additional experiments and analysis to study the effect of individual adaptation modules. 
Extensive experiments and ablation studies are conducted using multiple datasets, i.e., UCF-Sports, UCF-101 and JHMDB. 
Experimental results demonstrate the effectiveness of the proposed approach for addressing the domain shift of spatio-temporal action localization in multiple scenarios with domain discrepancies. 
\vspace{-10pt}

The contributions of this work are summarized as follows. First, we propose an end-to-end learning framework 
for solving the novel task of unsupervised domain adaptation in the context of spatio-temporal action localization. Second, we design and integrate three domain adaptation modules at the image-level (temporal and spatial) and instance-level (temporal) to alleviate the spatial and temporal domain discrepancy. Third, we propose a new experimental setup along with benchmark protocol and perform extensive adaptation experiments and ablation studies to analyze the effect of different adaptation modules and achieve state-of-the-art performance. Fourth, we demonstrate that not only does the individual adaptation of spatial and temporal features improve performance, but the adaptation is most effective when both spatial and temporal features are adapted.

\section{Related Work}
\subsection{Spatio-temporal Action Localization}
\label{spatiotemporal_related}
Spatio-temporal action localization is an active research topic in computer vision.  The goal is to localize and classify actions in both space and time. 
Majority of the existing approaches are supervised and can be categorized as either single frame or multi-frame. 
Most of the recent methods~\cite{gu2018ava,gkioxari2015finding,peng2016multi,saha2016deep,singh2017online,weinzaepfel2015learning} fall in the former category.
These schemes extend object detection frameworks~\cite{girshick2015fast,ren2015faster} to first generate region proposals and then classify them into actions at the frame level using a two-stream variant which processes both RGB and flow data separately. 
The backbone of these networks is generally a 3D CNN (e.g., C3D~\cite{tran2015learning} or I3D~\cite{carreira2017quo}). 
The resulting per-frame detections are then linked using dynamic programming~\cite{gkioxari2015finding,singh2017online} or tracking~\cite{weinzaepfel2015learning}. 
Some recent approaches, however, aim to jointly estimate localization and classification over several frames~\cite{kalogeiton2017action} or use 3D convolutions to predict short tubes~\cite{hou2017tube}. There has been recent attempts to learn without labels as well~\cite{soomro2017unsupervised}, where unlabeled data is used to automatically generate labels and train the classifiers.
%
%
%
\vspace{-10pt}
\subsection{Domain Adaptation}
\label{da_related}
Domain adaptation aims to bridge the gap between the source and target data collected from different domains.
%
Recent domain adaptation techniques under both semi-supervised and unsupervised settings have been introduced for image applications~\cite{csurka2017comprehensive}. 
The majority of these methods have been dedicated to applications involving image classification~\cite{saenko2010adapting,tzeng2014deep,long2015learning,ganin2015unsupervised,sener2016learning,motiian2017unified,haeusser2017associative}, object detection~\cite{chen2018domain,saito2019strong}, and semantic segmentation~\cite{sankaranarayanan2018learning,zou2018unsupervised,tsai2018learning}. 
Several unsupervised domain adaptation approaches use adversarial learning on the intermediate feature representations
to align the feature distribution between the two domains~\cite{ganin2015unsupervised,bousmalis2016domain,tzeng2017adversarial,chen2018domain}. 
%
%
%

In contrast, much less attention has been paid to adapt models for video analysis between domains, and especially for activity understanding.
%
%
While some progress has been made in this field recently, it is limited to whole-clip action recognition~\cite{jamal2018deep,chen2019temporal,pan2019adversarial} and action segmentation~\cite{chen2020action1,chen2020action2}.
One reason can be attributed to the fact that a well-organized setting to develop and benchmark the performance of domain adaptation algorithms for spatio-temporal action localization
does not exist. 
Existing datasets, e.g., CMU~\cite{ke2005efficient}, MSR Actions~\cite{yuan2009discriminative}, UCF Sports~\cite{rodriguez2008action}, and JHMDB~\cite{jhuang2013towards} provide spatio-temporal annotations but only for a small number of short video clips. 
The DALY~\cite{weinzaepfel2016human}, UCF-101~\cite{soomro2012ucf101} and AVA~\cite{gu2018ava} datasets address some of the aforementioned limitations by providing large-scale annotatios for spatio-temporal action localization. 
However, these datasets have very few overlapping categories amongst them. Additionally, the annotation setting of AVA is different from the other datasets, making it difficult to evaluate domain adaptation algorithms. 
%
%

To the best of our knowledge, this work is one of the first to adapt spatio-temporal action localization under the unsupervised setting.
To evaluate the new task, we propose a new experimental setup and evaluation protocol for future development.

\begin{figure*}[t!]
\centering
\includegraphics[width=0.95\textwidth,height=6.0cm]{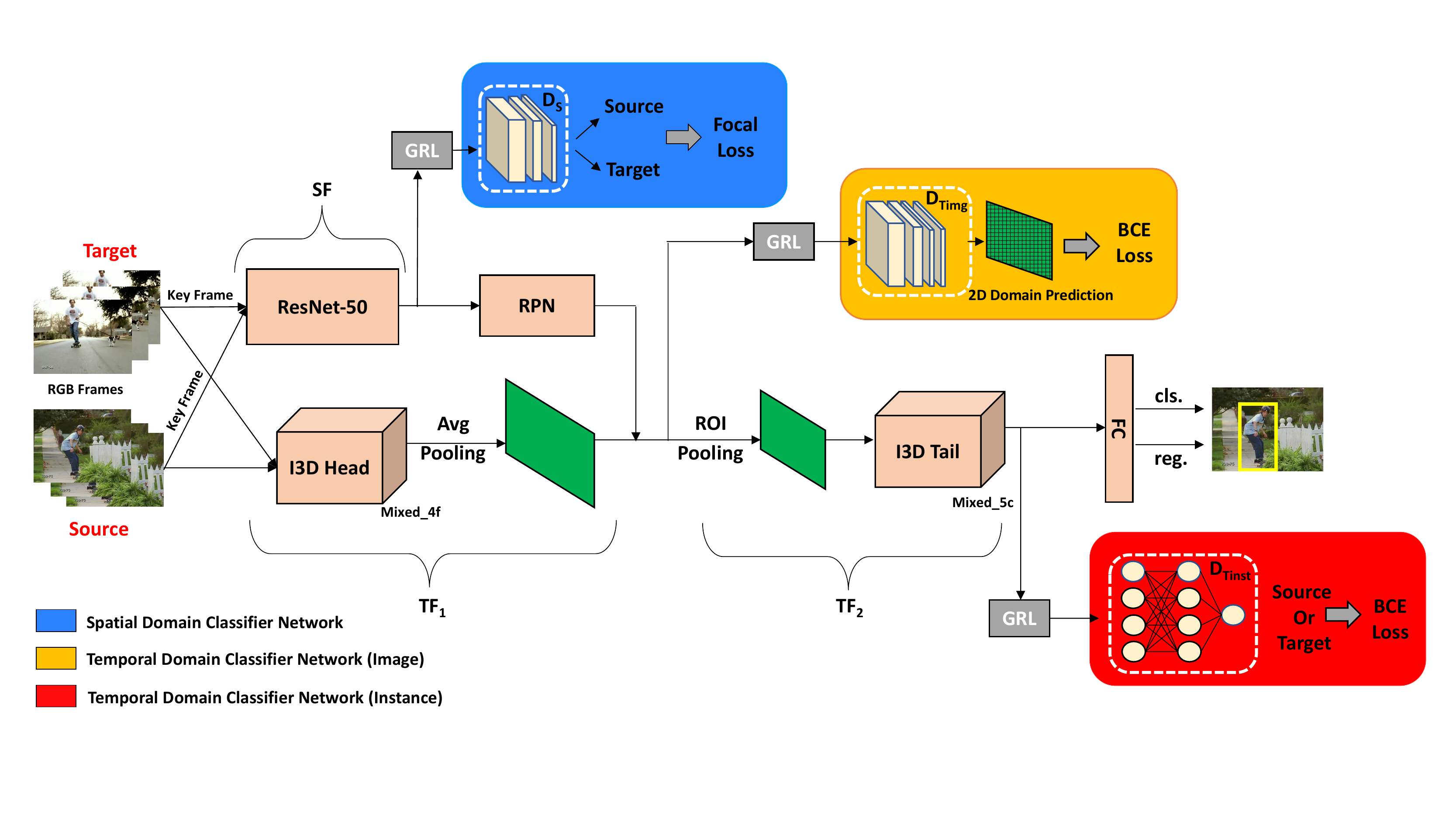}
\vspace{-10pt}
\caption{Proposed Network Architecture. The proposed algorithm aligns the distribution of both the spatial and temporal features of source and target domains for adapting actor proposals and action classification respectively. 
We use a spatial domain classifier network $D_s$ to align the spatial features generated by SF. 
The temporal features are adapted at the image and instance level using their respective temporal domain classifier networks, i.e., $D_{Timg}$ and $D_{Tinst}$. Image level features are extracted by TF\textsubscript{1} and instance level features are obtained from TF\textsubscript{2}.}\vspace{-10pt}
\label{fig:model}

\end{figure*}

\vspace{-15pt}
\section{Proposed Algorithm}
Our framework consists of an action localization model and three different adaptation modules for aligning both spatial and temporal feature distribution. The architecture of the proposed framework is shown in Figure~\ref{fig:model}.
%
%
%

\vspace{-10pt}
\subsection{Action Localization Model}
\label{basemodel}
%
Our model is based on the Faster R-CNN~\cite{ren2015faster} for end-to-end localization and classification of actions~\cite{peng2016multi}.
To model the temporal context, the I3D model~\cite{carreira2017quo} is incorporated.
%
%
The I3D model takes a video $V$ of length $T$ frames and generates the corresponding temporal feature representation using feature extractors $TF_1$ and $TF_2$ (see Fig.~\ref{fig:model}). 
Here, $TF_1$ extracts and temporally flattens the image level features from the fused mixed\_4f layer of I3D, which has a spatial and temporal stride of 16 pixels and 4 frames, respectively. %
%
This results in a compact representation of the entire input sequence. 
 
For the actor proposal generation, we use a 2D ResNet-50 model as the spatial encoder $SF$ (see Fig.~\ref{fig:model}) on the keyframe $K$ as the input for the region proposal network (RPN). 
We note $K$ is also the middle frame of an input clip to I3D.
The proposals are generated using the conv4 block of ResNet~\cite{he2016deep}. 
As the spatial stride of the conv4 block is also 16 pixels, we directly use the actor RPN proposals on $TF_1(V)$ and perform ROI pooling to obtain a fixed size representation of 7$\times$7$\times$832. 
This feature representation is then passed through $TF_2$, which uses the remaining I3D layers up to mixed\_5c and an average pooling layer to output an instance level feature vector of size 1$\times$1$\times$1024. 
%
This feature is used to learn an action classifier and a bounding box regressor. The loss function of the action localization model is formulated:
\begin{equation}
    \mathcal{L}_{act}= \mathcal{L}_{rpn} + \mathcal{L}_{cls} + \mathcal{L}_{reg},
\end{equation}
where $\mathcal{L}_{rpn}$, $\mathcal{L}_{cls}$, $\mathcal{L}_{reg}$ are the loss functions for the RPN, final classifier and box regressor respectively. 
The details regarding these individual loss functions can be found in the original paper~\cite{ren2015faster}.

\vspace{-5pt}
\subsection{Adaption in Space and Time}
\label{adaptation}
The adaptation process is comprised of two components: i) actor proposal adaptation and ii) action classification adaptation.
%

%
%
%
%

%
%
\vspace{1mm}
\noindent \textbf{Actor Proposal Adaptation.} 
We present a method based on adversarial learning to align the distribution of source and target features for the actor proposal network.
Specifically, the spatial domain discriminator $D_S$ is designed to discriminate whether the feature $SF(K)$ is from the source or the target domain.
Motivated by~\cite{saito2019strong}, the domain classifier is trained to ignore easy-to-classify examples and focus on hard-to-classify examples with respect to the classification of the domain by using the Focal Loss~\cite{lin2017focal}.
This prevents strong alignment between global features, which is both difficult and not desirable when there is a considerable domain shift. 
The loss is based on domain label $d$ of the input image, where $d=0$ refers to $K$ from the source domain and $d=1$ refers to $K$ from the target domain. 
The estimated probability by $D_S$ for the class with label $d = 1$ is denoted by $P\in [0,1]$, where $P$ is defined as:
\begin{equation}
P=
\begin{cases}
D_S(SF(K)), & \text{if } d=1,\\
1-D_S(SF(K)), & \text{otherwise.}
\end{cases}
\end{equation}
We formulate the spatial discriminator loss function as:  
\begin{align}
  \mathcal{L_{D_S}}  =  -\Big(&\frac{1}{n_s}\sum_{i=1}^{n_s} (1-P^s_i)^\gamma\text{log}(P^s_i)+  \frac{1}{n_t}\sum_{j=1}^{n_t} (P^t_j)^\gamma\text{log}(1-P^t_j)\Big),
\end{align}
where $n_s$ and $n_t$ denote the number of source and target samples in a minibatch respectively, and $\gamma$ controls the weight on hard to classify examples.

The gradient reversal layer (GRL)~\cite{ganin2016domain} is placed between the spatial domain discriminator $D_S$ and spatial feature extractor $SF$.
It helps $SF$ generate domain invariant features $SF(K)$ that fool the discriminator while $D_S$ tries to distinguish the domain.

\vspace{1mm}
\noindent \textbf{Action Classification Adaptation.} 
We extend adaptation in the case of images, specifically object detection~\cite{chen2018domain}, to videos by proposing to adapt the temporal features at both the image and instance level. While the former focuses on aligning global scene features that serve as context for actions, the latter reduces domain shift between the actor/action dynamics.   
Specifically, we use $TF_1$ as a feature extractor for adaptation at the image level and $TF_2$ for adaptation at the instance level. 
%
%
%
The $TF_1$ takes a video clip $V$ of $T$ frames and generates a compact feature representation $TF_1(V)$ using temporal pooling. 
We find that adaptation after temporal pooling of features performs well as although the actions in our experiments vary in terms of temporal dynamics across datasets, the datasets are not explicitly designed to capture that notion.
%
This characteristic is also shown in~\cite{chen2019temporal} for certain cases where adaptation after temporal pooling performs on par with explicit temporal adaptation modeling.    
The temporal domain discriminator $D_{Timg}$ then takes $TF_1(V)$ as input and outputs a 2D domain classification map $Q=D_{Timg}(TF_1(V))\in \mathbb{R}^{H\times W}$.
The parameters $H$ and $W$ are determined based on the resolution of $V$ as the spatial strides of $TF_1$ and $D_{Timg}$ are fixed. 
We then apply binary cross-entropy (BCE) loss on $Q$ based on the domain label $d$ of the input video $V$, where $d=0$ if $V$ belongs to the source domain, and $d=1$ if $V$ belongs to the target domain. 
The loss function for $D_{Timg}$ is formulated as:
\begin{align}
    \mathcal{L}_{D_{Timg}} = -\Big(&\frac{1}{n_s}\sum_{i=1}^{n_s} \sum_{h,w}(1-d_i)\text{ log }(1-Q_i^{(h,w)})+ \frac{1}{n_t}\sum_{j=1}^{n_t} \sum_{h,w}d_j\text{log}Q_j^{(h,w)}\Big),
\end{align}

where $h$ and $w$ correspond to the spatial indices of an activation in $Q$.

%
The instance level representation generated by $TF_2$ refers to the ROI-based feature vectors before they are fed to the final category classifiers (i.e., the FC layer in Figure~\ref{fig:model}). 
The instance level temporal domain classifier $D_{Tinst}$ takes the feature vector $TF_2(TF_1(V))$ as input and outputs a domain classification output for the $k$-th region proposal in the $i$-th image as $R_{i,k}$. 
The BCE loss is used to generate the final output.
The corresponding loss function is formulated as:
\begin{align}
    \mathcal{L}_{D_{Tinst}}=-\Big(&\frac{1}{n_s}\sum_{i=1}^{n_s}\sum_{k}(1-d_i)\text{ log }(1-R_{i,k}) +  \frac{1}{n_t}\sum_{j=1}^{n_t}\sum_{k} d_j\text{log}R_{j,k}\Big), 
\end{align}
    
where $d=0$ if $V$ belongs to the source distribution and $d=1$ if $V$ belongs to the target distribution.

\vspace{-10pt}

\subsection{Overall Objective}
The overall objective combines losses from the action localization model and the domain adaptation modules. 
We denote the overall adversarial loss from domain adaptation modules as:
\begin{equation}
    \mathcal{L}_{adv}(SF,TF,D)=\mathcal{L_{D_S}}+\mathcal{L}_{D_{Timg}}+\mathcal{L}_{D_{Tinst}}.
\end{equation}
%

For the adaptation task $s\rightarrow t$, given the source video $V^s$ and target video $V^t$, and by extension their corresponding key frames $K^s$ and $K^t$ respectively, the overall min-max loss function of the proposed framework is defined as the following:
\begin{align}
    \mathcal{L}(V^s,K^s,V^t,K^t)= \mathcal{L}_{act} + \lambda \mathcal{L}_{adv},
\end{align}
where $\lambda$ is a weight applied to the adversarial loss that balances the action localization loss.
 
\vspace{-10pt}
\section{Experiments and Analysis}
\label{exp}
We propose new experimental settings for developing and evaluating domain adaptation algorithms for spatio-temporal action localization as there is no existing benchmarks.
We first focus on the scenario of adapting from a smaller annotated domain to a much larger and diverse dataset.
%
and then provide some additional experiments and ablation studies to highlight the effect of the different adaptation modules used in the proposed approach.

The proposed approach is evaluated on three widely used benchmark datasets for action localization, namely UCF-101~\cite{soomro2012ucf101}, JHMDB~\cite{jhuang2013towards}, and UCF-Sports~\cite{rodriguez2008action}.
These datasets are gathered from different sources (suitable for domain adaptation evaluation) and are also commonly used for adaptation of action recognition~\cite{chen2019temporal,pan2019adversarial}. Additionally, their suitability for our experiments is further shown through the results where 
%
%
%
%
%
for each adaptation scenario, we show the baseline results of action localization (I3D+RPN) trained on the source data without applying domain adaptation, and a supervised model trained fully on the target domain data (oracle) to illustrate the existing domain shift between the datasets.
%
%
\vspace{-5pt}
\subsection{Datasets and Metrics}

\vspace{1mm}
\noindent \textbf{UCF Sports.} UCF Sports~\cite{rodriguez2008action} contains various trimmed sports actions collected from broadcast television channels.
It includes 10 actions, out of which we use 4 for our experiments which are common with UCF-101: \textit{Diving}, \textit{Golf-Swing}, \textit{Horse-Riding}, \textit{Skate-Boarding}. We use the train/test split as suggested in~\cite{lan2011discriminative}.

\vspace{1mm}
\noindent \textbf{UCF-101.} This action localization dataset~\cite{soomro2012ucf101} is purely collected from YouTube and contains more than 13000 videos and 101 classes.
We use 4 classes that are common with UCF-Sports from a 24-class subset with spatio-temporal annotations provided by~\cite{singh2017online}.
We conduct experiments on the official split 1 as is standard.

\vspace{1mm}
\noindent \textbf{JHMDB.} 
JHMDB~\cite{jhuang2013towards} is collected from sources ranging from digitized movies to YouTube, and consists of 928 trimmed clips over 21 classes. 
%
Each action class consists of varying number of clips (up to 40 frames).
We use the official split 1 for our experiments, and only use 3 classes which are common with UCF-101: \textit{Shoot Ball}, \textit{Golf}, \textit{Walk}. 

\vspace{1mm}
\noindent \textbf{Metrics.} We use the standard evaluation protocols and report intersection-over-union (IoU) performance using mean average precision (mAP) on both frame-level and video-level using an IOU threshold of 0.5.  
For frame-level IoU, the PASCAL VOC challenge protocol~\cite{everingham2015pascal} is used. 
%
For video-level IoU, we follow~\cite{peng2016multi} to form action tubes by linking frame-level detections using dynamic programming and calculate 3D IoUs.
%
%

%

\noindent \textbf{Implementation Details.}
We implement the proposed algorithm in Pytorch. 
ResNet-50 and I3D networks are initialized with pre-trained models based on ImageNet~\cite{deng2009imagenet} and Kinetics~\cite{kay2017kinetics} datasets, respectively. 
%
%
%
%
%
%
%
For the proposed adaptation method, we first pre-train the action localization network using the source domain clips, and then fine-tune the network for adaptation.
We use different adaptation networks for each of the adaptation modules.
%
%
%
%
%
%
%
%
%
%
%
%
%
More experimental details and results can be found in the supplementary material. 
%
The source code and trained models will be made available to the public. 

\begin{table}[t]
\centering
\caption{\label{tab:ucfsports_ucf101} Frame and video mAP results for adaptation from UCF-Sports to UCF-101 \textit{with} (left) and \textit{without} (right) background frames.
}\vspace{10pt}
\begin{subtable}
\small
\centering
\resizebox{.45\columnwidth}{!}{
\setlength{\tabcolsep}{3pt}
\begin{tabular}{c lll llll l l}
\toprule[1pt]
Method & \multicolumn{1}{c}{\begin{tabular}[c]{@{}c@{}}T\\ img\end{tabular}} & \multicolumn{1}{c}{\begin{tabular}[c]{@{}c@{}}T\\ ins\end{tabular}} & \multicolumn{1}{c}{\begin{tabular}[c]{@{}c@{}}S\\ img\end{tabular}} & \multicolumn{1}{c}{\begin{tabular}[c]{@{}c@{}}Div\\ ing\end{tabular}} & \multicolumn{1}{c}{\begin{tabular}[c]{@{}c@{}}Glf\\ Swg\end{tabular}} & \multicolumn{1}{c}{\begin{tabular}[c]{@{}c@{}}Hrs\\ Rdg\end{tabular}} & \multicolumn{1}{l}{\begin{tabular}[c]{@{}l@{}}Skt\\ Bdg\end{tabular}} & \multicolumn{1}{c}{\begin{tabular}[c]{@{}c@{}}Fr.\\ mAP\end{tabular}} & \multicolumn{1}{c}{\begin{tabular}[c]{@{}c@{}}Vid.\\ mAP\end{tabular}} \\ 
\midrule[.5pt]
I3D+RPN &  &  &  & 7.1 & 56.3 & 30.7 & 39.5 & 33.4 & 57.1  \\ \hline
\multirow{4}{*}{Ours} & \checkmark &  &  & 12.2 & 64.6 & 40.0 & 41.9 & 39.7 & 61.0  \\ 
 &  & \checkmark &  & 12.2 & 64.9 & 40.7 & 42.3 & 40.0 & 61.6  \\ 
 &  &  & \checkmark & 13.9 & 64.9 & 51.5 & 51.8 & 45.5 & 68.9  \\ 
 & \checkmark &  & \checkmark & 14.9 & 64.1 & 56.2 & 54.9 & 47.5 & 70.6  \\ 
 &  & \checkmark & \checkmark & 13.0 & \textbf{68.8} & 51.3 & 50.6 & 45.9 & 67.1  \\ 
 & \checkmark & \checkmark & \checkmark & \textbf{17.9} & 63.3 & \textbf{63.0} & \textbf{55.0} & \textbf{49.8} & \textbf{73.6} \\ \hline
Oracle &  &  &  & 90.4 & 97.6 & 94.2 & 91.0 & 93.3 & 99.0 \\
\bottomrule[1pt]
\end{tabular}}
\end{subtable}
\hspace{2em}
\begin{subtable}
\small
\centering
\resizebox{.45\columnwidth}{!}{
\setlength{\tabcolsep}{3pt}
\begin{tabular}{c lll llll l l}
\toprule[1pt]
Method & \multicolumn{1}{c}{\begin{tabular}[c]{@{}c@{}}T\\ img\end{tabular}} & \multicolumn{1}{c}{\begin{tabular}[c]{@{}c@{}}T\\ ins\end{tabular}} & \multicolumn{1}{c}{\begin{tabular}[c]{@{}c@{}}S\\ img\end{tabular}} & \multicolumn{1}{c}{\begin{tabular}[c]{@{}c@{}}Div\\ ing\end{tabular}} & \multicolumn{1}{c}{\begin{tabular}[c]{@{}c@{}}Glf\\ Swg\end{tabular}} & \multicolumn{1}{c}{\begin{tabular}[c]{@{}c@{}}Hrs\\ Rdg\end{tabular}} & \multicolumn{1}{l}{\begin{tabular}[c]{@{}l@{}}Skt\\ Bdg\end{tabular}} & \multicolumn{1}{c}{\begin{tabular}[c]{@{}c@{}}Fr.\\ mAP\end{tabular}} & \multicolumn{1}{c}{\begin{tabular}[c]{@{}c@{}}Vid.\\ mAP\end{tabular}} \\ 
\midrule[.5pt]
I3D+RPN &  &  &  & 6.9 & 44.7 & 30.2 & 39.0 & 30.2 & 18.1  \\ \hline
\multirow{6}{*}{Ours} & \checkmark &  &  & 11.7 & 51.0 & 39.3 & 41.6 & 35.9 & 22.6  \\ 
 &  & \checkmark &  & 11.6 & 51.1 & 40.0 & 42.1 & 36.2 & 22.5  \\ 
 &  &  & \checkmark & 13.3 & 50.9 & 50.8 & 51.5 & 41.7 & 22.3  \\ 
 & \checkmark &  & \checkmark & 14.2 & 51.1 & 55.5 & 54.6 & 43.8 & 24.0  \\ 
 &  & \checkmark & \checkmark & 12.4 & \textbf{53.7} & 50.5 & 50.3 & 41.7 & 21.6  \\ 
 & \checkmark & \checkmark & \checkmark & \textbf{16.9} & 51.8 & \textbf{62.2} & \textbf{54.7} & \textbf{46.4} & \textbf{24.1} \\ \hline
Oracle &  &  &  & 83.2 & 67.9 & 92.8 & 91.0 & 83.7 & 56.6 \\
\bottomrule[1pt]
\end{tabular}}
\end{subtable}
\vspace{-15pt}
\end{table}



\vspace{-10pt}
\subsection{Adaptation to Large-Scale Data}
Adapting a model learned from a small dataset to a large unlabeled domain is more challenging than typical settings in the literature, and is also more useful as annotating large amount of data is infeasible for spatio-temporal action localization. 
%
%
%
In this work, the target domain is UCF-101, and the sources are UCF-Sports and JHMDB sets. 
Note that the source datasets are much smaller in size and less diverse than the target one, details of which can be found in the supplementary material.

%

\vspace{1mm}
\noindent \textbf{UCF-Sports $\rightarrow$ UCF-101.} 
%
We conduct experiments on the common classes from both the datasets and show the results in 
%
%
%
%
Table~\ref{tab:ucfsports_ucf101}. Since UCF-101 is an untrimmed set, we show results both with and without considering background frames, the latter also requiring temporal localization. Note that we do not use background frames during training, making the latter setting extremely challenging. 

\begin{figure}[t]
\centering
\includegraphics[width=.8\textwidth,height=4.0cm]{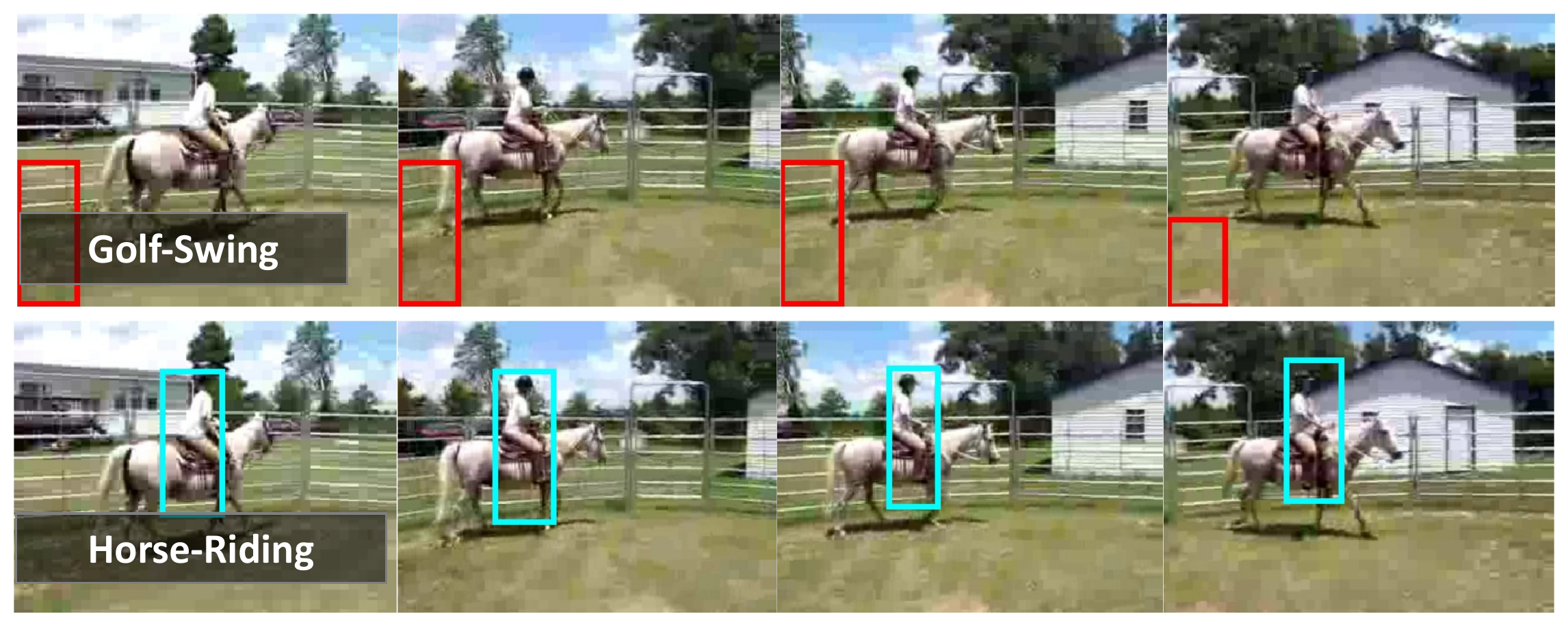}\\
\caption{Example clip of \textit{Horse-Riding} action from UCF-101, with  
baseline model (\textcolor{red}{red}) and our best adapted model (\textcolor{cyan}{cyan}) shown and predicted label overlaid. \vspace{-15pt}
}
\label{ucfsports_ucf101_fig}
\end{figure}

Although UCF-Sports is also a sport-oriented dataset like UCF-101, a significant performance gap between the baseline and oracle results is observed, suggesting significant domain shift and difficulty for adaptation.
%
 %
For aligning temporal features, both image level as well as instance level adaptation yield similar and considerable improvement over the baseline of 6.3\% and 6.6\% for frame-mAP, and 3.9\% and 4.5\% for video-mAP respectively, as shown in Table~\ref{tab:ucfsports_ucf101} (left). However, alignment of spatial features, which is responsible for adapting the actor proposals yields 12.1\% (frame-mAP) and 11.8\% (video-mAP) improvement.
The results demonstrate the importance of localizing the action in space, as it is necessary to localize the action first before classification. 
Finally, we show that the combination of aligning both spatial and temporal features leads to the best results, with performance gains of 16.4\% (frame-mAP) and 16.5\% (video-mAP).  
%
Note that the improvement also generalizes well across different categories, suggesting the effectiveness of the proposed framework in reducing domain discrepancy. Figure~\ref{ucfsports_ucf101_fig} shows an example from the UCF-101 dataset, where the baseline model (without adaptation) fails to detect \textit{Horse-Riding} action while the adapted model correctly localizes and classifies the action.
%
 
When the background frames are considered in Table~\ref{tab:ucfsports_ucf101} (right), we observe similar trends after adding the adaptation modules. The absolute numbers however, are lower, indicating the presence of false positives on the background frames. Although this does not drastically affect the frame-mAP, the video-mAP is considerably affected. 
This also suggests that an explicit mechanism should be developed to handle background frames during adaptation, especially for temporal localization.

\begin{table}[b]
\centering
\vspace{-15pt}
\caption{\label{tab:jhmdb_ucf101} Frame and video mAP results for adaptation from JHMDB to UCF-101 (left) and UCF-101 to JHMDB (right). 
}
\vspace{10pt}
\begin{subtable}
\small
\centering
\resizebox{.45\columnwidth}{!}{
\setlength{\tabcolsep}{4pt}
\begin{tabular}{c lll lll l l}
\toprule[1pt]
Method & \multicolumn{1}{c}{\begin{tabular}[c]{@{}c@{}}T\\ img\end{tabular}} & \multicolumn{1}{c}{\begin{tabular}[c]{@{}c@{}}T\\ ins\end{tabular}} & \multicolumn{1}{c}{\begin{tabular}[c]{@{}c@{}}S\\ img\end{tabular}} & \multicolumn{1}{c}{\begin{tabular}[c]{@{}c@{}}Golf\\ Swg\end{tabular}} & \multicolumn{1}{c}{\begin{tabular}[c]{@{}c@{}}Bskt\\ Ball\end{tabular}} & \multicolumn{1}{c}{Walk} & \multicolumn{1}{c}{\begin{tabular}[c]{@{}c@{}}Fr.\\ mAP\end{tabular}} & \multicolumn{1}{c}{\begin{tabular}[c]{@{}c@{}}Vid.\\ mAP\end{tabular}} \\ 
\midrule[.5pt]
I3D+RPN &  &  &  & 62.6 & 38.2 & 47.2 & 49.3 & 51.8 \\ \hline
\multirow{4}{*}{Ours} & \checkmark &  &  & 64.3 & 40.8 & 50.6 & 51.9 & 56.4 \\ 
 &  & \checkmark &  & 64.5 & 40.8 & 50.8 & 52.0 & 56.7 \\ 
 &  &  & \checkmark & 74.5 & 56.9 & 55.3 & 62.2 & 69.0 \\ 
 & \checkmark &  & \checkmark & 73.7 & 56.9 & 54.4 & 61.7 & 64.1 \\ 
 &  & \checkmark & \checkmark & 73.8 & 58.6 & 55.5 & 62.6 & 68.2 \\ 
 & \checkmark & \checkmark & \checkmark & \textbf{75.1} & \textbf{59.2} & \textbf{56.2} & \textbf{63.5} & \textbf{69.5} \\ \hline
Oracle &  &  &  & 95.7 & 87.0 & 90.4 & 91.0 & 88.2 \\
\bottomrule[1pt]
\end{tabular}}
\end{subtable}
\begin{subtable}
\small
\centering
\resizebox{.45\columnwidth}{!}{
\setlength{\tabcolsep}{4pt}
\begin{tabular}{c lll lll l l}
\toprule[1pt]
Method & \multicolumn{1}{c}{\begin{tabular}[c]{@{}c@{}}T\\ img\end{tabular}} & \multicolumn{1}{c}{\begin{tabular}[c]{@{}c@{}}T\\ ins\end{tabular}} & \multicolumn{1}{c}{\begin{tabular}[c]{@{}c@{}}S\\ img\end{tabular}} & \multicolumn{1}{c}{\begin{tabular}[c]{@{}c@{}}Golf\\ Swg\end{tabular}} & \multicolumn{1}{c}{\begin{tabular}[c]{@{}c@{}}Bskt\\ Ball\end{tabular}} & \multicolumn{1}{c}{Walk} & \multicolumn{1}{c}{\begin{tabular}[c]{@{}c@{}}Fr.\\ mAP\end{tabular}} & \multicolumn{1}{c}{\begin{tabular}[c]{@{}c@{}}Vid.\\ mAP\end{tabular}} \\ 
\midrule[.5pt]
I3D+RPN &  &  &  & 86.6 & 27.2 & 38.4 & 50.7 & 60.7  \\ \hline
\multirow{4}{*}{Ours} & \checkmark &  &  & 88.5 & 36.3 & 42.9 & 55.9 & 68.7  \\ 
 &  & \checkmark &  & 87.1 & 35.4 & 42.9  & 55.1 & 71.7  \\ 
 &  &  & \checkmark & 94.9 & 35.6 & 55.4 & 62.0 & 71.0  \\ 
 & \checkmark & \checkmark & \checkmark & \textbf{96.4} & \textbf{46.7} & \textbf{57.9} & \textbf{67.0} & \textbf{75.4} \\ \hline
Oracle &  &  &  & 96.6 & 70.5 & 87.0 & 84.7 & 93.4 \\
\bottomrule[1pt]
\end{tabular}}
\end{subtable}
\end{table}

\noindent \textbf{JHMDB $\rightarrow$ UCF-101.}
%
%
While UCF-101 is comprised of activities in the sports domain, JHMDB consists of videos from everyday activities (some sport-related sequences are also included). 
%
Note that from the set of common classes, {\em walk} action in JHMDB is visually very different from the {\em walking with dog} action in UCF-101. 
However, we still incorporate the {\em walk} action in our experiments to increase the number of common classes.
%
%
%
%
%
We show the results in Table~\ref{tab:jhmdb_ucf101} (left) without considering background frames, but still consider temporal localization for \textit{Walk} action as it has few sequences containing multiple action instances. 
The performance gap between baseline and oracle results suggests a significant domain shift.
%
A considerable improvement is obtained by adaptation of either spatial or temporal features for both frame and video mAPs, and their combination leads to the best performance gain of 14.2\% (frame-mAP) and 17.7\% (video-mAP) over the baseline.

We also observe that differently from~\cite{chen2018domain}, instance level feature alignment combined individually with image level spatial feature adaptation does not yield much improvement and performs worse in some cases. This is because~\cite{chen2018domain} focuses only on spatial feature alignment from the same backbone at image level before RPN and instance level before classification, while we are dealing with both temporal and spatial feature alignment from two separate backbones (i.e., I3D and Resnet-50). Consequently, as shown in the Table~\ref{tab:jhmdb_ucf101} (left) and Table~\ref{tab:ucfsports_ucf101}, temporal feature adaptation at image level is needed, which highlights the importance of our design choice -- adaptation for both spatial (image level) and temporal (image and instance level) features.
%
%
%
%
%
%
The results also suggest that both spatial context and actor/action dynamics are equally important for action classification, as both types of temporal features are required for best performance and yield similar improvement over the baseline.
%
%
%
%

\vspace{-10pt}
\subsection{Additional Experiments and Analysis}
%
%
In this section, we study the effect of adapting from a larger annotated domain to a much smaller dataset.
%
%
We discuss the empirical results and analyze the effects of the individual adaptation modules by studying the classification and localization errors of the different models.

\vspace{1mm}
\noindent \textbf{UCF101 $\rightarrow$ JHMDB.} We use UCF-101 and JHMDB as the source and target datasets respectively, with the same set of common classes as before.  
%
%
\begin{figure}[t]
\centering
\includegraphics[width=.95\textwidth,height=3.0cm]{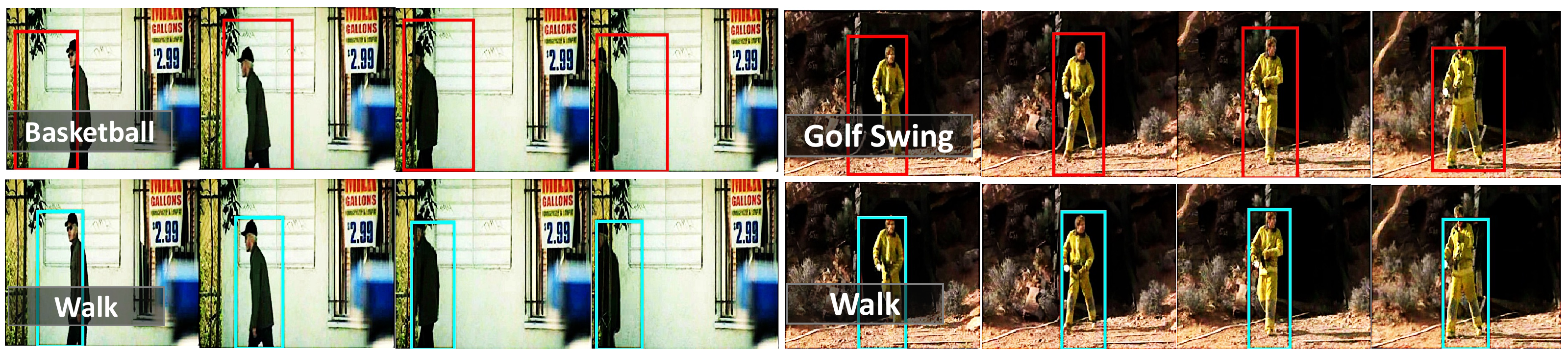}\\
\caption{Example clips of \textit{Walk} action from JHMDB, with  
baseline model (\textcolor{red}{red}) and our best adapted model (\textcolor{cyan}{cyan}) shown and predicted label overlaid. \vspace{-10pt}
}
\label{walk_comp}
\end{figure}
Even when adapting from a much larger database to a smaller dataset, we observe similar trends in Table~\ref{tab:jhmdb_ucf101} (right) as before, with the significant gap between the baseline and oracle results suggesting that even having large amount of annotations does not help much in the case of domain shift.
Note that the domain gap mainly comes from two classes: \textit{Basketball} and \textit{Walk}. 
The baseline performance for \textit{Golf-Swing} is very close to the oracle results due to a significant amount of training labels in UCF-101.
%
%
However, while {\em Walk} in UCF-101 contains about 20 times more samples than in JHMDB, the baseline performance is far from the oracle result because of the significant visual differences of the action between the datasets.
%
%
Specifically, \textit{Walk} action in UCF-101 is always accompanied with a dog in outdoor environments. 
Due to this, the model trained on UCF-101 (without adaptation) finds it hard to classify \textit{Walk} action on JHMDB, as shown in Figure~\ref{walk_comp}. 
Adaptation helps alleviate visual differences and improves localization performance. Examples of visual differences can be found in the supplementary material.

%

%
%
%

%
%
%
%
%
%

\begin{figure}[t!]
\centering
\includegraphics[width=0.95\textwidth,height=3.0cm]{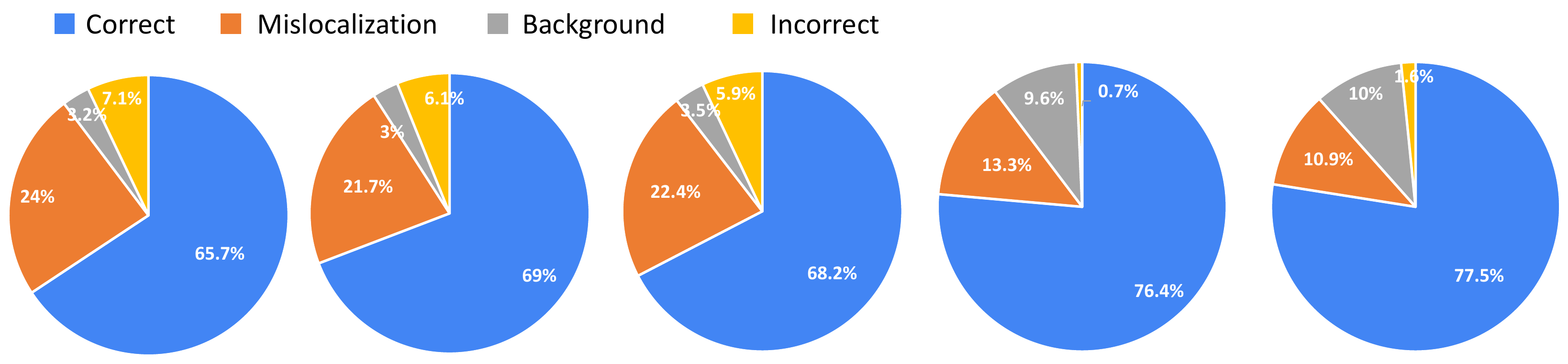}\\
\hspace{.1cm}\small{(a) I3D+RPN} \hspace{.4cm}(b) Ours (T. Img) \hspace{.5cm}(c) Ours (T. Ins) \hspace{.5cm}(d) Ours (Spatial) \hspace{.5cm}(e) Ours (all)\\
\caption{Error analysis of top ranked detections. Fraction of predictions that are correct, mislocalized, are confused with background or incorrectly predicted are shown.}\vspace{-10pt}
\label{fig:error}
\end{figure}

\noindent \textbf{Error Analysis on Top Ranked Detections.} To study the effect of the individual adaptation modules, we analyze the classification and localization errors from the most confident detections of the model.

We use the UCF-101 $\rightarrow$ JHMDB experiment for analysis. 
Since the JHMDB dataset is a small set, we select the top 1000 predictions based on the corresponding predicted confidence score by the baseline model (i.e., I3D+RPN) and our models with various adaptation modules. 
Motivated by~\cite{hoiem2012diagnosing,mettes2019pointly}, we categorize the detections into four error types: i) \textbf{correct}: the detection has an
overlap $\in [0.5,1]$ with the ground-truth; ii) \textbf{mis-localized}:
the detection has an overlap $\in [0.3,0.5)$; iii) \textbf{background}: the detection has an overlap $\in [0.0,0.3)$, which means it takes a background as a false positive; and iv) \textbf{incorrect}: the detection has a different class than the ground truth. 
The first three errors are related to the localization error given the detected class is correct, while the last error measures the incorrect classifications. 
In addition, we also analyze the errors of the bottom 1000 detections in the supplementary material, with the goal to understand the extent of the adaptation effect. 

Figure~\ref{fig:error} shows that temporal feature alignment at both image and instance level improves the correct detections as well as reduces the mislocalized error. 
It also reduces the incorrect classifications. 
The spatial feature alignment, in addition to increasing the correct detections, also considerably reduces the mislocalized error. 
This can be attributed to that spatial features directly improve the RPN, which is responsible for actor proposal generation. 
It also reduces the incorrect classification. 
In addition, we note that there is an increase in the background error, which can be considered as duplicate detections as these are not incorrectly classified. 
However, as expected, our model with both spatial and temporal features aligned increases the correct detections the most and also gives the least mislocalization error. 

\vspace{-15pt}
\section{Conclusion and Future Work}
In this paper, we propose a new task and an end-to-end approach for unsupervised domain adaptation for spatio-temporal action localization. Our approach is built by extending the Faster R-CNN algorithm. 
In order to reduce domain shift, we design and integrate three domain adaptation modules at the image level (temporal and spatial) and instance level (temporal). Experimental results demonstrate that significant performance gain can be achieved when spatial and temporal features are adapted separately, or jointly for the most effective results.
%

Our experimental setup lacks in large number of overlapping classes and significant temporal variations between the datasets (mentioned in Section~\ref{adaptation}), which is a restriction of the problem space as there does not exist such datasets. Our work is an essential first step to stimulate the community to collectively build large-scale benchmark datasets and algorithms for domain adaptation of spatio-temporal action localization.
%
%
%

\noindent \textbf{Acknowledgment.} This work is supported in part by the NSF CAREER Grant \#1149783 and Honda Research Institute USA.

\bibliography{egbib}
\end{document}